\def\year{2019}\relax
\documentclass[letterpaper]{article} %DO NOT CHANGE THIS
\usepackage{aaai19}  %Required
\usepackage{times}  %Required
\usepackage{helvet}  %Required
\usepackage{courier}  %Required
\usepackage{url}  %Required
\usepackage{graphicx}  %Required
\usepackage{multirow}
\usepackage{color}
\usepackage{bm}
\usepackage{diagbox}
\usepackage{cancel} 
\usepackage{mathrsfs} 
\usepackage{eqnarray}
\usepackage{times}
\usepackage{epsfig}
\usepackage{graphicx}
\usepackage{amsmath}
\usepackage{amssymb}
\usepackage{caption}
\usepackage{subcaption}
\newcommand{\floor}[1]{\lfloor #1 \rfloor}

\usepackage[utf8]{inputenc} % allow utf-8 input
\usepackage[T1]{fontenc}    % use 8-bit T1 fonts
\usepackage[citecolor=blue,linkcolor=blue,urlcolor=red]{hyperref}       % hyperlinks
\usepackage{booktabs}       % professional-quality tables
\usepackage{amsfonts}       % blackboard math symbols
\usepackage{nicefrac}       % compact symbols for 1/2, etc.
\usepackage{microtype}      % microtypography

\begin{document}
\title{Single-Label Multi-Class Image Classification by
	Deep Logistic Regression}

 \author{
 	Qi Dong,\textsuperscript{\rm 1}
 	Xiatian Zhu,\textsuperscript{\rm 2}
 	Shaogang Gong\textsuperscript{\rm 1}\\
 	\textsuperscript{\rm 1}Queen Mary University of London,
 	\textsuperscript{\rm 2}Vision Semantics Ltd.\\
 	q.dong@qmul.ac.uk, eddy@visionsemantics.com, s.gong@qmul.ac.uk
 }
 
\maketitle
\begin{abstract}
 The objective learning formulation is essential for the success of 
 convolutional neural networks.
 In this work, we analyse thoroughly the standard learning objective functions
 for multi-class classification CNNs: softmax regression (SR) for {\em single-label} scenario and logistic regression (LR) 
 for {\em multi-label} scenario.
 Our analyses lead to an inspiration of exploiting LR
 for single-label classification learning,
 and then the disclosing of the {\em negative class distraction} problem in LR.
 To address this problem, we develop two novel LR based objective functions
 that not only generalise the conventional LR 
but importantly turn out to be competitive alternatives to SR in single label classification.
 Extensive comparative evaluations demonstrate
 the model learning advantages of the proposed LR functions
 over the commonly adopted SR
 in single-label coarse-grained object categorisation and 
 cross-class fine-grained person instance identification tasks.
 We also show the performance superiority of our method
 on clothing attribute classification in comparison
 to the vanilla LR function. 
 {\em The code had been made publicly available}.
 % at \url{https://github.com/qd301/FocusRectificationLogisticRegression}.
\end{abstract}

\section{Introduction}
\label{introduction}
Convolutional neural networks (CNNs) \cite{lecun1989backpropagation}
have demonstrated impressive performance success in 
a wide variety of image recognition problems 
\cite{krizhevsky2012imagenet,he2016deep,liu2016deepfashion}.
Two common problems are 
{\em single-label} (one class label per image)
and 
{\em multi-label} (multiple class labels per image)
classification.
Whilst both problems have the {\em same learning objective} of 
inducing a multi-class classifier CNN model through supervised training,
their standard objective learning functions are {\em rather different}.

Specifically, in single-label classification learning,
we often adopt the \texttt{softmax regression} (SR)
learning algorithm.
This is based on the per-sample single-label and class-exclusion assumptions \cite{bridle1990probabilistic}.
For multi-label classification
in which a data sample may be tied with multiple class labels, 
we instead adopt 
the \texttt{logistic regression} (LR) learning algorithm
\cite{Bishop}.
Without the ``single-label'' and ``class-exclusion'' prior,
LR considers per-sample prediction of all individual class labels
{\em independently}.

It is intuitive that single-label classification
is a special case of multi-label classification.
Hence, LR should be applicable 
for single-label classification. Surprisingly, despite that both SR and LR 
have been extensively studied and exploited
for learning single-label and multi-label classification {\em independently}, their comparison in deep learning of single-label classification has never been investigated systematically in the literature to our knowledge. 
A few fundamental questions remain unclear: 
How advantageous is the {\em de facto standard} choice SR 
over LR for single-label classification on earth? 
Is LR possibly competitive with or even superior to SR?

In this work, we investigate the potential and validity of 
{\em the Logistic Regression learning algorithm
	for single-label multi-class classification}
in theory and practice. 
We observe that 
although SR has an advantage 
of conducting class discriminative learning 
via posing a competition mechanism between
the ground-truth and other classes per training sample,
it may simultaneously distort the underlying data manifold
geometry which may in turn hurt the model generalisation capability
\cite{belkin2006manifold}.
This is because in SR
all non-ground-truth classes are identically pushed away
from the labelled ground-truth class in a homogeneous fashion
with the class-to-class correlation ignored in model learning.
On the contrary, LR avoids this limitation 
due to that each class is modelled independently
as a separate binary classification task \cite{Bishop}.
This allows the intrinsic inter-class geometrical correlation
to emerge naturally in a data-mining manner.

In light of the above theoretical merit, 
we particularly study the efficacy of LR for single-label 
classification learning.
Empirically, we found that
the vanilla LR indeed yields less discriminative and generalisable
models 
on image classification tasks in most cases.
With in-depth LR loss and gradient analysis,
we identify that the {\em negative class distraction} problem turns out to be
the major model learning barrier.

We make three {\bf contributions}:
{\bf (1)} 
We investigate the fundamental characteristics of SR and LR in single-label multi-class classification learning.
This is an essential and under-studied problem 
in the literature to our knowledge.
{\bf (2)} 
We identify the negative class distraction problem in LR
as the main obstacle for single-label classification learning, 
and propose two optimisation focus rectified LR learning algorithms in 
a hard mining principle as effective solvers.
{\bf (3)} 
We conduct extensive evaluations on two single-label multi-class classification based image recognition tasks, coarse-grained object categorisation and fine-grained zero-shot person instance identification, by using {\em five} standard benchmarks.
The results show that our methods perform on par with or outperform the standard 
algorithm SR.
We further validate the effectiveness of our method
in clothing attribute classification with extremely sparse labels per data sample.

\section{Related Work}
\label{sec:rel}
With the recent surge of interest in 
neural networks like CNNs,
image classification by deep learning
has gained massive attention and remarkable success
\cite{krizhevsky2012imagenet, 
	girshick2015fast,dong2017multi,liu2016deepfashion,lin2017focal,dong2017class}.
We have witnessed significant advances in many aspects
including 
network architecture improvement \cite{he2016deep},  
nonlinear activations \cite{maas2013rectifier},  
layer designs \cite{lin2013network},
regularisation techniques \cite{srivastava2014dropout}, 
optimisation algorithms \cite{kingma2014adam},  
and data augmentation \cite{krizhevsky2012imagenet}.

Essentially, these existing methods are mostly used and
grounded on the well-established learning algorithms
such as softmax regression (SR)
\cite{luce2005individual,bridle1990probabilistic} 
and
logistic regression (LR)
\cite{little1974existence,mor1990ranking}.
Impacted by the traditional design principles \cite{goodfellow2016deep,krishnapuram2005sparse}, LR is often used to produce the prediction output
of multi-label multi-class classification models \cite{liu2016deepfashion,chua2009nus},
whilst SR to that of single-label multi-class classification models
\cite{krizhevsky2012imagenet,russakovsky2015imagenet} in current deep learning practice.

Although taking {\em different} learning formulations, 
both SR and LR algorithms aim to train a multi-class neural network classifier
which, once trained, is able to predict the top one or multiple
class label(s) of new samples at test time.
One reason for this design discrepancy is that in SR,
the single-label constraint 
facilitates the learning of a multi-class classifier. 
Although lacking the single-label class prior,
LR has a merit of individually learning per-class distributions
and better maintaining the class manifold structures
\cite{Bishop}.
In spite of that, LR is however {\em rarely} chosen to 
learn single-label multi-class
classification by existing methods, 
leaving its potential efficacy for image recognition remaining unknown in 
deep learning.
To fill this gap, we systematically study this ignored problem,
identify and address a negative class distraction (NCD) problem.

The NCD problem is concerned with imbalance learning with 
a particular focus on positive and negative classes per training sample.
It is therefore related to 
the conventional class imbalanced learning problem \cite{japkowicz2002class,weiss2004mining,he2009learning,dong2018imbalanced,huang2016learning}.
Whilst sharing some theoretical concept in general,
the NCD problem in our context is fundamentally different 
because it is {\em independent} of the {\em training data distribution} which
however is the core problem existing class imbalanced learning methods aim to address.
Differently, the NCD problem is underpinned in the {\em target class} space,
occurring in the multi-class joint optimisation process on each training sample. 
A larger class space leads to a more serious NCD problem.
In other words, the NCD problem remains 
even with {\em absolutely balanced} (equally sized) training samples 
per class.
Besides, LR bias correction has been extensively studied
\cite{king2001logistic,schaefer1983bias,qiu2013logistic,heinze2002solution} but 
still focusing on the data imbalance issue,
rather than the NCD problem as considered in this work.

\section{Delve Deep into Deep Learning Classification}
\label{sec:loss_analysis}
{Supervised deep learning algorithms} learn to classify input images into target class labels,
given a training set of $n$ image-label pairs 
$\mathcal{D}=\{(\bm{x}_{i}, {y}_{i})\}_{i=1}^{n}$
where ${y}_{i} \subseteq \mathcal{Y} = \{1,2,\cdots,K\}$ specifies 
the ground-truth label set of image $\bm{x}_{i}$
with one (single-label)
or multiple (multi-label) class(es) associated.
There are totally $K$ possible classes. 
Supervised learning of such multi-class classifiers is generally conducted based on 
estimating probabilistic class distributions ${\bm{p}}$ of training images with
the element $p_k = p(k \; | \; \bm{x})$, $k \in \mathcal{Y}$.

\begin{figure*}[th] 
	\centering
	\includegraphics[width=0.9\linewidth]{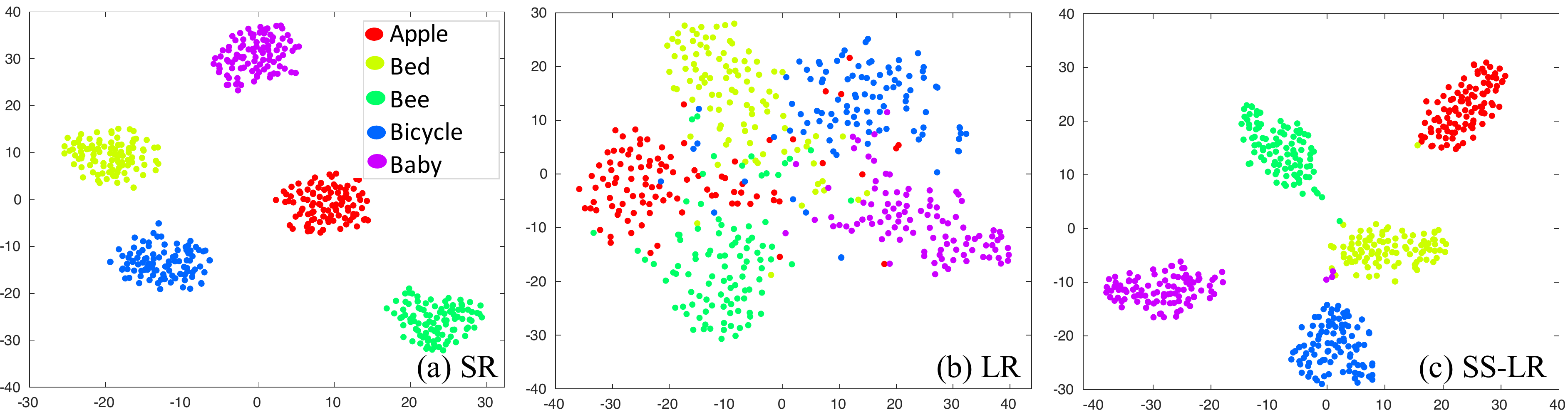} 
	\caption{
		Colour-coded t-SNE feature embedding of 5 CIFAR100 object classes  
		produced by 
		(a) SR,
		(b) LR,
		and 
		(c) the proposed SS-LR (Eq \eqref{eq:LR_att})
		loss functions. It is evident that   
		in the feature embedding space by the vanilla LR,
		different classes are poorly distinguishable with severe cross-class 
		boundary overlap. 
		In contrast, our SS-LR 
		yields more discriminative feature embedding   
		by addressing the NCD problem involved in model optimisation.
		Best viewed in colour. 
	}
	\label{fig:vis}
\end{figure*}

\noindent{\bf Probability Distribution Estimation. }
To make $p_k$ represent valid probability values, 
{one} common approach to normalising individual $p_k$ is to 
apply the logistic (or sigmoid) function \cite{little1974existence,mor1990ranking}  
to squash the raw output 
$\bm{z} = \phi (\bm{x} \;|\;\bm{\theta})$
into the interval $(0,1)$ as: 
\begin{equation}\small
\begin{aligned}
p_k = p(k \; | \; \bm{x};\bm{\theta})
= 
\sigma(\bm{z})_k = 
\frac{1}{1+e^{-z_k}} \\
= \frac{e^{z_k}}{e^{z_k}+1},
\quad k \in \mathcal{Y} = \{1,2,\cdots,K\}
\label{eq:sigmoid}
\end{aligned}
\end{equation}
where $\bm{\theta}$ denotes the model parameters  
that project an input $\bm{x}$ into a logit space $\bm{z}$
via a to-be-learned mapping $\phi$.

In essence, Eq \eqref{eq:sigmoid} models a Bernoulli distribution for each individual class (a binary-value random variable) 
{\em independently},
i.e. the model learns to predict the positive probability $p(1 \; | \: \bm{x})$ for each class $k$.
Therefore, it naturally fits the {\em multi-label} classification scenario:
Each image sample can be associated with multiple (an unknown number) class labels
in any possible combinatorial ways.
Eq \eqref{eq:sigmoid} is known as \texttt{Logistic Regression} (LR).

{\em Single-label} classification is another common scenario
in which only one class label is outputted for a single sample.
This implicitly assumes a mutual exclusion relationship between all classes.
Hence, we can further require the {\em entire} vector $\bm{p}$ as a multi-class probability distribution: 
$\sum_{k=1}^K p_k = 1$ and $p_k \geq 0$.
To that end, 
the softmax function is often employed \cite{luce2005individual,peterson1989new,bridle1990probabilistic}.
Formally, we exponentiate
and normalise the logit $\bm{z}$ to obtain a valid probability vector ${\bm{p}}$ as:
\begin{equation}
\begin{aligned}
p_k = p(k \; | \; \bm{x};\bm{\theta})
=  
h(\bm{z})_k 
= \frac{e^{z_k}}{\sum_{j=1}^K e^{z_j}}, \\
\text{with} \;\; k \in \mathcal{Y} = \{1,2,\cdots,K\}
\end{aligned}
\label{eq:softmax}
\end{equation}
Eq \eqref{eq:softmax} models a  
categorical  
 distribution of a discrete variable over
multi-classes {\em collectively} and {\em inter-dependently}.
Eq \eqref{eq:softmax} is called \texttt{Softmax Regression} (SR).

\noindent {\bf Learning Objective Function. }
To perform supervised classification learning, 
we usually employ the principle of maximum likelihood
that attempts to match the model distribution with 
the data empirical distribution by a cross-entropy measurement
\cite{goodfellow2016deep}.
The specific learning objective function
relies on the regression form of the model' s prediction.

In case of multi-label classification,
the objective function for maximum likelihood learning is formulated as:
\begin{equation}
\footnotesize
\mathcal{L}_{\text{LR}}(\bm{x}, y)  = - \sum_{k=1}^{K}
\Big (q_{k} \log(p_{k}) + 
(1-q_{k})\log (1-p_{k}) \Big )
\label{eq:LR}
\end{equation}
This objective function aggregates the 
negative log-likelihood of all class-wise Bernoulli distributions.

In case of single-label classification, 
we directly use the cross-entropy between 
the ground-truth class distribution $\bm{q}$ of the training datum 
and the predicted class distribution $\bm{p}$ of the model
to form the objective function.
$\bm{q} \!=\! \delta_{k,y}$ is Dirac delta which equals to 1
if $k\!=\!y$, and 0 otherwise. 
The learning objective function 
is written as:
\begin{equation}
\mathcal{L}_\text{SR}(\bm{x}, y) = - \sum_{k=1}^{K}
q_k \log p_k =-\log {p_y}
\label{eq:SM}
\end{equation}

\noindent {\bf Remarks. } 
In essence, the key of model learning 
is  
to induce the target multi-class feature embedding space. 
A {\em generalisable} feature space should be characterised by 
an accurate  
inter-class manifold structure.
Given a training sample $\bm{x}$, SR enforces a {\em competition} 
between the ground-truth and other classes
to learn the model discrimination capability:
the softmax output always sums to 1, subject to that an increase in the estimation of one class necessarily corresponds to a decrease in the estimation of others.
Whilst this competition significantly helps learn discriminative inter-class boundaries, it may distort the underlying inter-class manifold structure
therefore potentially hurting the model generalisation capability \cite{belkin2006manifold},
since SR treats all non-ground-truth classes {\em identically}
by pushing them away from the ground-truth class in a homogeneous manner.
In contrast, LR learns to induce the inter-class manifold structure 
from the training data, enabling a natural emergence of the underlying multi-class manifold geometry.

\begin{figure*}[th] 
	\centering
	\includegraphics[width=1\linewidth]{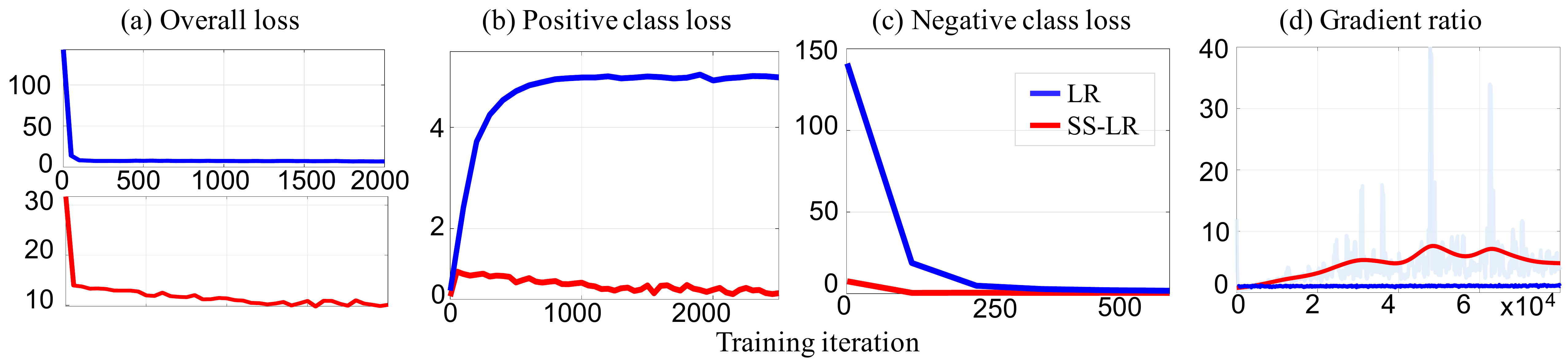}
	\caption{
		Negative class distraction effect of LR
		on Tiny ImageNet.  .
		(a) Overall training loss values;
		The loss of (b) the positive class and (c) the negative classes;
		(d) Gradient ratio of positive to negative classes.
	}
	\label{fig:obs}
\end{figure*}

\section{
	Focus Rectified Logistic Regression}
\label{sec:method}

\subsection{The Negative Class Distraction Problem}

{Despite the theoretical merit of LR,
	directly using the vanilla LR often leads to 
	inferior model performance than SR, as shown in Fig \ref{fig:vis} (b).
	Why does this happen?
	To examine this problem in LR, 
	we track and analyse the training loss and gradient quantities (the blue curves in Fig \ref{fig:obs}).
	We observe that at the beginning of model training, 
	the LR loss drops dramatically until near 0 (see Fig \ref{fig:obs} (a)). 
	Further decomposing the LR loss into two parts: 
	the positive class loss on the ground truth class (see Fig \ref{fig:obs}(b)) 
	and negative class loss on all non-ground-truth classes (see Fig \ref{fig:obs}(c)),
	we find that at early training iterations,
	(1) the starting negative class loss is far larger than the starting positive class loss  (e.g. 140 vs 0.4); and  
	(2) the positive class loss increases {\em unexpectedly}, while the negative class loss drops fast. 
	This suggests that at the early training stage, the overall LR loss is dominated by negative classes therefore the positive class is largely ignored.
	We call this effect as {\em negative class distraction} (NCD).

	The NCD problem is intrinsic to single-label multi-class classification.
	Specifically, suppose a $K$-class setting, 
	each training sample $\bm{x}$ has only one positive class (the ground-truth label $y$) 
	but $(K\!-\!1)$ negative classes. 
	With the vanilla LR learning objective (Eq \eqref{eq:LR}), the positive class obtains insufficient attention, especially when $K$ is large. 
	Therefore, the training is hinted by the severe learning bias towards the negative classes. 
	Such a biased loss composition deceives the model to converge towards some poor local minima with the negative classes well satisfied 
	(Fig \ref{fig:obs}(c)) 
	whilst the positive class largely ignored (Fig \ref{fig:obs}(b)). 
	This can be further justified by the nearly zero gradient ratio of positive to negative classes  
	 (Fig \ref{fig:obs})(d)). 
	 The NCD problem similarly exists in multi-label classification with sparse labels per sample (Fig \ref{fig:losstrack}).}

{Often, the inherent learning difficulty and velocity of different classes 
	can be distinct, as indicated in the accuracy variety over object classes \cite{deng2009imagenet}.
	Hence, treating all negative classes per sample identically as 
	Eq \eqref{eq:LR} may be {\em not} optimal,
	and selecting important ones to learn is likely to be more effective.
	Crucially, this helps mitigate the NCD problem
	as more learning focus is assigned to the positive class.
	Inspired by such consideration,
	we formulate two negative class selection mechanisms 
	to rectify the biased learning focus of the vanilla LR
	in a hard mining principle:
	A training sample $\bm{x}$ is more (less)
	informative to hard (easy) classes.}

\subsection{Rectification by Negative Class Hard Selection}

We confine the learning focus to non-trivial (hard) other than all negative classes.
Specifically, we use the predicted probability $p_k$ as the 
hardness measurement to rank $(K\!-\!1)$ negative classes in the descending order.
We then choose the top negative classes to formulate the objective loss function as:
\begin{equation} 
\mathcal{L}_{\text{LR}}^\text{hs}(\bm{x}, y)  = 
- \log(p_y) 
- \alpha \sum_{k \in f_{hs}(m|\bm{p}, y)}
\Big(\log(1 - p_{k}) \Big)
\label{eq:LR_sel}
\end{equation}
where the hard selection function $f_{hs}(m|\bm{p}, y)$ returns the
top $m\%$ negative classes with highest prediction probabilities.
We add a balancing weight $\alpha$ to trade-off 
positive and negative classes, inspired by
the cost-sensitive learning \cite{akbani2004applying,he2009learning}.

Adjusting $m \in [0, 100]$ allows us to modulate the focus rectification degree:
with a training sample, we learn the decision boundaries of 
$m\%$ most confusing negative classes along with the positive class. 
We empirically found that $m\!=\!25$ is satisfactory.
When $m\!=\!100$, we attend all negative classes
with a cost-sensitive trade-off between the positive and all negative classes.
The weight $\alpha$ can be intuitively set as inversely proportional
to the selected negative class number: $\alpha\!=\!\frac{\beta}{\floor{m\% (K-1)}} < 1$ 
where $\beta$ is a hyper-parameter ($\beta\!=\!10$ in our experiments).
We call this {\em negative class Hard Selection} based LR formulation as \textbf{\em HS-LR}.

\subsection{Rectification by Negative Class Soft Selection}

An alternative to HS-LR is a soft selection of negative classes. Formally, we employ a hardness (probability) adaptive weight $(p_{k})^r$ to each negative class as:
\begin{equation} 
\mathcal{L}_{\text{LR}}^\text{ss}(\bm x)  = 
- \log(p_y) 
- \alpha \sum_{k=1, k \neq y}^{K}
\Big( (p_{k})^r \log(1 - p_{k}) \Big)
\label{eq:LR_att}
\end{equation}
where $r \geq 0$ is the attending parameter
that controls the rate of attending hard {negative} classes
and disregarding easy {negative} classes.
{Note that} when $r\!=\!0$, it is equivalent to HS at $m\!=\!100$.
Growing $r$ makes the {focus}
modulating effect {like-wisely} increase.
In our experiments, we found that $r$ is not sensitive {in a reasonable range of } and $r\!=\!2$ {is selected in our main experiments}
(Fig \ref{fig:gamma}).
We set $\alpha\!=\!\frac{\beta}{K-1}$ since all negative classes
are considered.

Our soft selection mechanism achieves the effect of hard class mining 
in this manner:
If a negative class $k$ is a hard 
class w.r.t. $\bm{x}$ 
and receives a higher probability $p_k$,
its weight $(p_{k})^r$ is larger and hence more attention is assigned.
When the value of $p_k$ is small 
which suggests an easy negative class, 
the learning attention will be close to 0 and the quantity 
of class $k$ is significantly down-weighted.
We call this {\em negative class Soft Selection} based LR formulation as \textbf{\em SS-LR}.

The soft selection
principle has been used in other
methods, e.g. Entropy-SGD \cite{chaudhari2016entropy}
and focal loss \cite{lin2017focal}.
Entropy-SGD tackles a different problem 
of seeking better local minima. 
The focal loss is more similar to 
our SS-LR (Eq \eqref{eq:LR_att}) but differs in a number of
fundamental aspects:
(1) Focal loss solves the {\em global} training data sample imbalance
whilst SS-LR deals with the {\em local} sample-wise negative class distraction,
independent of the training data distribution over classes.
(2) Focal loss is built on the softmax regression,
whilst SS-LR is formulated based on the logistic regression.
(3) Focal loss aims to suppress easy training instances in the {\em sample} space
whilst SS-LR handles the per-sample negative classes in the {\em class} space.

\begin{figure}[th] 
	\centering
	\includegraphics[width=0.9\linewidth]{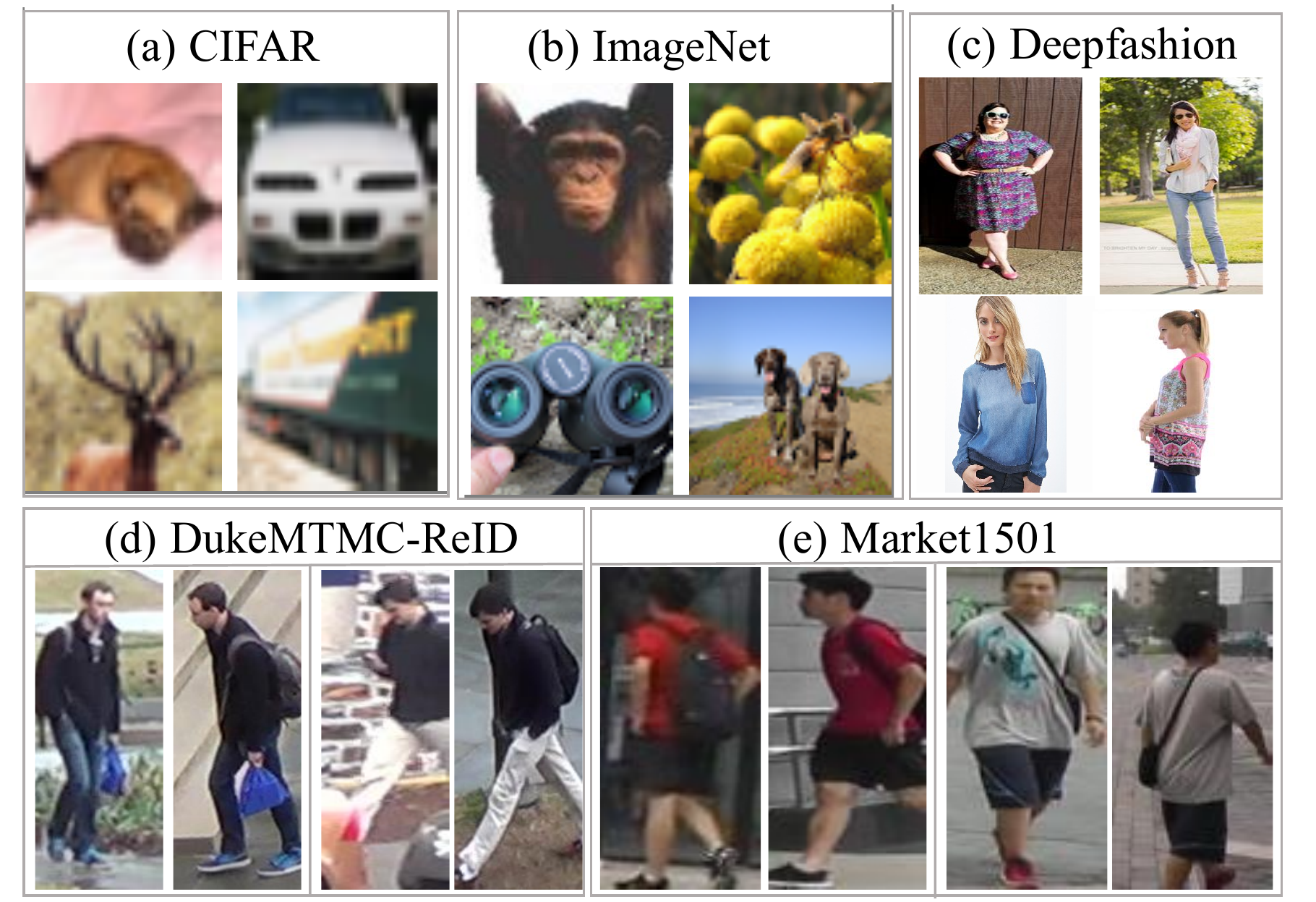}
	\caption{
		Example images of benchmark datasets evaluated.
	}
	\label{fig:datasets}
\end{figure}

\section{Experiments}
\label{sec:exp}

We evaluated the proposed method on 
(1) single-label object classification,
(2) person instance identification, and 
(3) multi-label clothing attribute recognition.
For each test, we performed 10 independent runs and reported the average result.
Note that, 
outperforming existing best performers  
by extra complementary techniques   
is not the focus of our evaluations. 
Rather, the key is to investigate the model generalisation performance 
of the same network model learned by
the SR and LR objective functions 
in a fair test setting.

\subsection{Single-Label Object Image Classification}
\label{sec:exp_obj}
\noindent {\bf Datasets. }
We used three single-label object classification benchmarks.
\textbf{\em\small CIFAR10} and \textbf{\em\small CIFAR100} \cite{krizhevsky2009learning}
both have $32\!\times\!32$ sized images from 10 and 100 classes, respectively.
We adopted the benchmarking 50,000/10,000 train/test image split on both.
\textbf{\em\small Tiny ImageNet (Tiny200)} \cite{deng2009imagenet} 
contains
110,000 64$\times$64 images from 200 classes.
We followed the standard 100,000/10,000 train/val setting. 
These datasets present varying class numbers,
thus giving a spectrum of different single-label model test scenarios.
Example images of these datasets are given in Fig \ref{fig:datasets}.

\noindent{\bf Experiment setup. }
We carried out all the following experiments in TensorFlow \cite{abadi2016tensorflow}.
We tested {three} varying-capacity networks: ResNet-32 
(32 layers with 0.7 million parameters) \cite{he2016deep}, 
WideResNet-28-10 
(28 layers with 36.5 million parameters)
\cite{zagoruyko2016wide},
and DenseNet-201 (20 million parameters) \cite{huang2017densely}.
We adopted the top-1 classification accuracy in our evaluations. 
We used the standard SGD with momentum for model training.
We set 
the initial learning rate to 0.1,
the momentum to 0.9,
the weight decay to $10^{-4}$,
the batch size to 128/64{/128} for CIFAR/Tiny200{/ImageNet},
the epoch number to 300.
We set the parameter $m$ (Eq \eqref{eq:LR_sel}) 
in the range of $[25,75]$ and
$r=2$ (Eq \eqref{eq:LR_att}) ($r=50$ for ImageNet) by a grid search on the validation dataset.
Data augmentation includes horizontal flipping and translation.
All models compared used the same training/test data
for fair comparative evaluations.

\begin{table}[th]
	\centering
	\caption{
		Evaluation on single-label object image classification. 
		{Metric}: Top-1 accuracy rate (\%). 
	}
	\label{tab:obj_clf}
	\begin{tabular}{r||c |c| c}
		\hline
		Base Net & 
		\multicolumn{3}{c}{{\em ResNet-32} \cite{he2016deep}}
		\\
		\hline
		Dataset &{CIFAR10} & {CIFAR100}  &{Tiny200} \\\hline  
		SR 
		& 92.5    & 68.1    & 50.2 \\
		\hline
		LR          
		&\bf 93.0     & 64.3    & 45.9 \\  
		\hline
		
		\bf HS-LR
		&\bf 93.0 & 68.9 & 50.7 
		\\ \hline
		\bf SS-LR 
		&\bf 93.0 &\bf 69.1 &\bf  56.0
		\\ 
		\hline \hline
		
		Base Net &
		\multicolumn{3}{c}{{\em WideResNet-28-10}  \cite{zagoruyko2016wide}}
		\\
		\hline
		Dataset & {CIFAR10} & {CIFAR100}  &{Tiny200} \\\hline  
		SR 
		& 95.3      & 81.0     & 57.4      \\ 
		\hline
		LR 
		& 95.0      & 79.0    & 55.2     \\ 
		\hline
		\bf HS-LR
		& 95.3 &79.0 &59.0 
		\\ \hline
		\bf SS-LR
		&\bf 96.0  &\bf 81.2 &\bf  61.0      \\
		\hline
	\end{tabular}
\end{table}

\noindent{\bf Evaluation. } 
Table \ref{tab:obj_clf} compares the single-label object 
categorisation performances
between the SR and LR function variants
using the small ResNet-32 and large WideResNet-28-10 architectures.
We make these observations:
{\bf (1)} When the class number increases, 
the vanilla LR suffers a more severe NCD problem
and yields much weaker performances than SR.
For example, on CIFAR10 with 10 classes, 
LR performs on a par or even slightly better.
However, LR is clearly inferior on Tiny ImageNet with 200 classes
(On the other hand, this also simultaneously implies a good potential of LR
since the results are not far worse).
{\bf (2)}
The proposed LR variants
notably improve the performance  
and outperform the SR,
especially on tasks with more classes.
This indicates that once the NCD problem is properly solved (Fig \ref{fig:obs}),
LR can be a stronger formulation
for single-label classification learning.
This observation is rarely made in the literature
where SR dominants the learning of single-label classification models. 
{\bf (3)}
The Soft Selection (SS) strategy consistently yields
the best model generalisation,
suggesting the advantages of exploiting all negative object classes
in a hardness adaptive manner.
{\bf (4)}
Both small and large nets benefit from the proposed \text{LR} algorithms,
indicating that our method is generically applicable to 
different CNN architectures.

\subsection{Fine-Grained Person Instance Identification}
\label{sec:exp_reid}

\noindent {\bf Datasets. }
We used two popular person instance identification ({a.k.a.,} person re-identification) benchmark datasets in our experiments.
The \textbf{\em\small Market-1501} \cite{zheng2015scalable}
has 32,668 images of 1,501 different identities (ID) 
captured from 6 outdoor camera views.  
We followed the standard 751/750 train/test ID split.
The \textbf{\em\small DukeMTMC} \cite{ristani2016MTMC}
consists of 36,411 images of 1,404 IDs from 8 camera views.
We adopted the benchmarking 702/702 ID split as \cite{zheng2017unlabeled}.
Unlike normal image classification, 
person re-identification (re-id) is a more fine-grained
recognition problem of matching person instances across non-overlapping camera views. It is more challenging due to the inherent zero-shot learning knowledge transfer from seen classes (IDs) to unseen classes in deployment, 
i.e. no overlap between training and test classes.

\noindent{\bf Experiment setup. }
We tested two nets, with variant capacities, often used in existing re-id methods:
ResNet-50 \cite{he2016deep} 
(50 layers with 25.6 million parameters),
and MobileNet \cite{howard2017mobilenets}
(28 layers with 3.3 million parameters).
We adopted two standard performance metrics in the {\em single query} mode:
the Cumulative Matching Characteristic accuracy (Rank-$1$ rate)
and mean Average Precision (mAP).
We used the Adam optimiser \cite{kingma2014adam}, and 
set the initial learning rate to 0.0003,
the momentum to 0.9,
the weight decay to $10^{-4}$,
the batch size to 32, and 
the maximum epoch number to 300.
We trained all methods without using
complex tricks in order to focus the evaluation on 
comparing SR and LR algorithms.

\begin{table}[h!]
	\centering
	\caption{Evaluation on person instance identification.}
	\label{tab:reid}
	\begin{tabular}{r|cc|cc}
		\hline
		Base Net & 
		\multicolumn{4}{c}{{\em ResNet-50} \cite{he2016deep}}
		\\
		\hline
		Dataset
		& \multicolumn{2}{c|}{Market-1501} 
		& \multicolumn{2}{c}{DukeMTMC} 
		\\ \hline 
		Metric (\%)   
		& Rank-1 & mAP    & Rank-1 & mAP 
		\\ \hline 
		SR      
		& 83.3 & 65.8   
		& 73.7 & 54.9 \\
		\hline
		LR      
		&81.4   & 65.0   
		&  72.2  &  54.6  
		\\ 
		\hline 
		\bf HS-LR &\bf 87.1&\bf 70.7&\bf 77.9&\bf 60.1  \\
		
		\hline
		\bf SS-LR 
		& 85.8  & 69.7 
		& 76.7   &  58.2  \\
		\hline 
		\hline

		Base Net &	\multicolumn{4}{c}{{\em MobileNet} \cite{howard2017mobilenets}}
		\\
		\hline
		Dataset
		& \multicolumn{2}{c|}{Market-1501} 
		& \multicolumn{2}{c}{DukeMTMC} \\ \hline 
		Metric (\%)   
		& Rank-1 & mAP    & Rank-1 & mAP
		\\ \hline
		SR & 71.7  & 50.0    
		& 57.0&35.8     
		\\ \hline
		LR & 51.5  & 34.3   
		& 43.9  & 27.5      
		\\ \hline
		\bf HS-LR 
		&\bf 76.4 &\bf 54.1 &\bf 63.7   &\bf 42.5       
		\\ \hline
		\bf SS-LR
		& 74.0 & 53.7 &62.9  &  41.5         
		\\ \hline
	\end{tabular}
\end{table}

\noindent{\bf Evaluation. }
Table \ref{tab:reid} shows the performance comparisons of SR 
and LR methods on person re-id. 
We have the following observations:
{\bf (1)}
Unlike generic object classification (Table \ref{tab:obj_clf}), 
the vanilla LR and SR produce very 
similar generalisation performances when using ResNet-50.
The plausible reason is that, re-id has the less stringent requirement 
of well fitting the model to training classes
since the test classes are entirely new and unseen to model training.
{\bf (2)}
Both proposed LR algorithms
improve the model performance,
suggesting that the NCD problem still matters
in cross-class recognition.
{\bf (3)}
Hard selection (HS) turns out to be the best strategy,
as opposite to object classification (Table \ref{tab:obj_clf})
where SS is the best performer.
This indicates that
using every training sample to 
learn all classes is not necessarily superior,
which may negatively affect the modelling capacity of mining 
fine-grained discriminative information
among a large number of training classes (751 on Market-1501, and 702 on DukeMTMC).

To validate the statistical significance of our model's performance,
we conducted a Wilcoxon signed-rank test on the DukeMTMC results using MobileNet.
The test verifies that the improvements in accuracy and mAP rates are statistically significant at the 5\% significance level.

\subsection{Clothing Attributes Recognition}
\label{sec:exp_attr}

Apart from single-label classification, 
we evaluated our LR methods
on the multi-label classification specially with only a few labels per instance, which also suffers 
a similar NCD problem.
The SR is not applicable in this test.

\noindent {\bf Dataset. }
We evaluated a large scale multi-label clothing attribute dataset
\textbf{\em \small DeepFashion} \cite{liu2016deepfashion}.
This dataset has 289,222 images labelled with 1,000 fine-grained clothing attributes
with a 209,222/40,000/40,000 train/val/test benchmark setting. 
Each image is associated with {\em extremely} sparse labels, $3$ out of 1,000 in average.
The training set is also highly class imbalanced (733:1),
therefore presenting a very challenging 
multi-label classification task. 
We adopted the standard 
multi-label classification setting
without using auxiliary types of supervision such as 
key-points and clothing category 
as used in \cite{liu2016deepfashion}.

\begin{table}[!h]
	\centering
	\caption{Evaluation on multi-label attribute recognition.}
	\label{tab:df_balance}
	\begin{tabular}{r|cc|cc}
		\hline
		Base Net & 
		\multicolumn{4}{c}{{\em ResNet-50} \cite{he2016deep}}
		\\
		\hline
		\multirow{2}{*}{Metric (\%)} 
		& \multicolumn{2}{c|}{Accuracy} 
		& \multicolumn{2}{c}{mAP} 
		\\ \cline{2-5} 
		& Per Img    & Per Cls & Per Img    & Per Cls    
		\\ \hline  
		LR  
		& 64.8 & 50.1  & 21.6  & 2.5  
		\\ \hline  
		
		\bf HS-LR   
		&\bf74.3 &\bf59.0 &31.5 &\bf9.3
		\\ \hline
		
		\bf SS-LR   
		&73.9 & 58.7 &\bf34.4 &9.2   
		\\ 
		\hline  
		\hline

		Base Net & 
		\multicolumn{4}{c}{{\em MobileNet} \cite{howard2017mobilenets}}
		\\ \hline
		\multirow{2}{*}{Metric (\%)} 
		& \multicolumn{2}{c|}{Accuracy} 
		& \multicolumn{2}{c}{mAP} 
		\\ \cline{2-5} 
		& Per Img    & Per Cls & Per Img    & Per Cls    
		\\ \hline 
		LR & 62.4&51.9&23.4&4.7
		\\ \hline
		\bf HS-LR &\bf 72.2 &\bf57.4&28.1&\bf7.2 
		\\ \hline
		\bf SS-LR & 72.0&55.8&\bf31.0& 6.5       
		\\  \hline
	\end{tabular}
\end{table}

\noindent{\bf Experiment setup. }
We similarly tested two nets:
ResNet-50 \cite{he2016deep}, and
MobileNet \cite{howard2017mobilenets}.
We adopted two standard performance measurement criteria:
mean Average Precision (mAP)
and balanced classification accuracy \cite{dong2018imbalanced,huang2016learning}.
The latter is particularly designed
to remedy the performance evaluation bias towards
the majority classes of imbalanced data.
For each metric, we evaluated per-image and per-class model performances
of top-5 class predictions. 
We used the Adam optimiser \cite{kingma2014adam},
with 
the learning rate of 0.0001 {for the first 45 epochs and 0.00001 for the last 5 epochs,}
the weight decay of 0.00004, 
the momentum of 0.9,
and the batch size of 32.

\begin{figure}[h]
	\centering
	\includegraphics[width=0.9\linewidth]{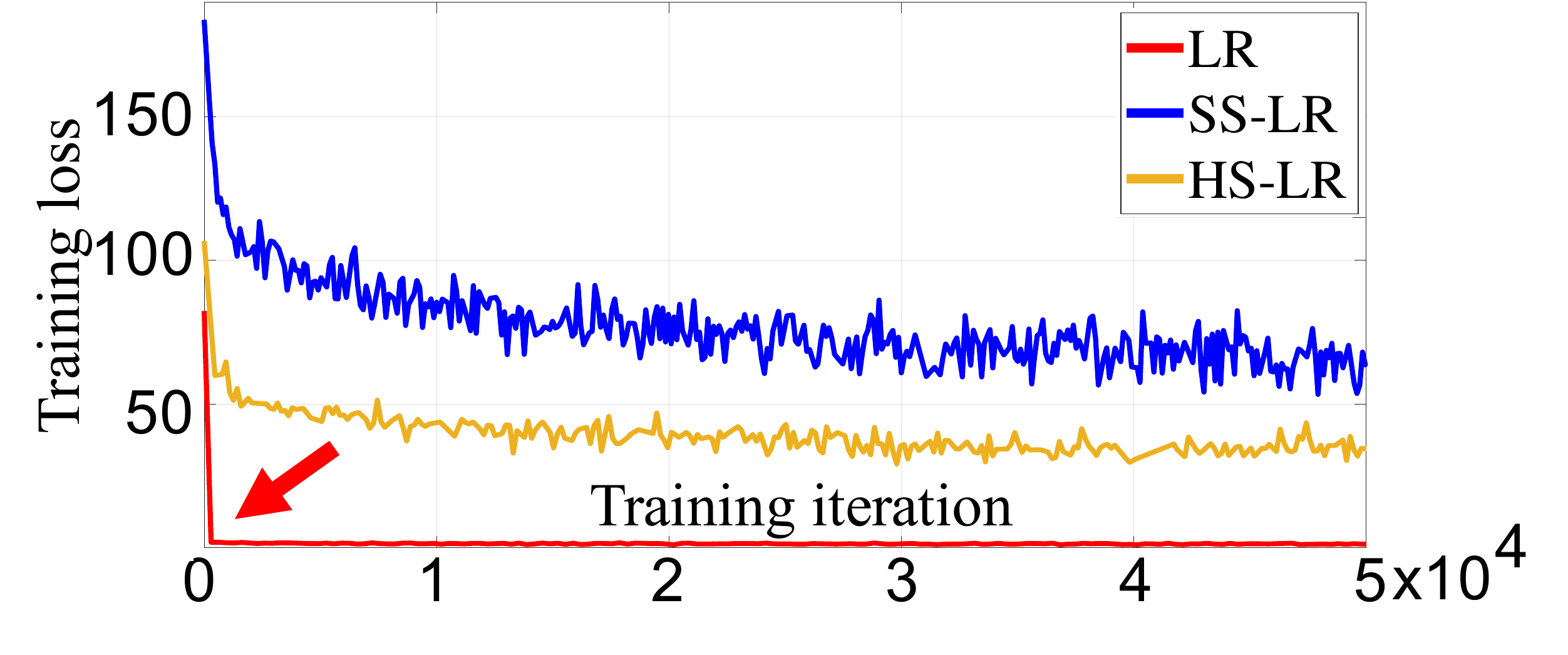}
	\captionof{figure}{Training loss on DeepFashion. }
	\label{fig:losstrack}
\end{figure}

\noindent{\bf Evaluation. }
Table \ref{tab:df_balance} shows the clothing attribute classification performances 
of different LR variants.
It is observed that:
{\bf (1)}
Our LR methods are significantly superior 
to the vanilla algorithm, which is consistent with 
the results on single-label object classification
and person re-id. 
{\bf (2)}
Hard and soft selection strategies perform
similarly across different nets and metrics.
These results suggest the generic advantages 
of our approach in multi-label classification,
confirming the existence of NCD.
Moreover, our method also notably outperforms
the state-of-the-art result (54.5 per-class accuracy)
in the same test setting obtained by \cite{dong2018imbalanced},
further validating the efficacy of our approach. 

Figure \ref{fig:losstrack} shows the
loss converging process during training.
Similar to single label object classification (Fig \ref{fig:obs}),
the vanilla LR is clearly hurt by the NCD problem.
In contrast, SS-LR and HS-LR achieve a more stable and healthy model learning process
by adaptively hard mining negative classes.

\subsection{Further Analysis and Discussion}
\label{sec:ablation}

{\bf Learning Focus Rectification. }
We employ the gradient ratio of positive to negative classes to explicitly reveal the model learning focus (the higher ratio, the more focus on the positive class and vice versa). By comparing two lines in Fig \ref{fig:obs} (d), it is clear 
that with the vanilla LR,
the positive class is highly distracted by
negative classes throughout model training.
The proposed SS-LR formulation effectively raises the 
learning focus of positive classes in training
and therefore mitigating the NCD problem.

\begin{figure}[h]
	\centering
	\includegraphics[width=0.8\linewidth]{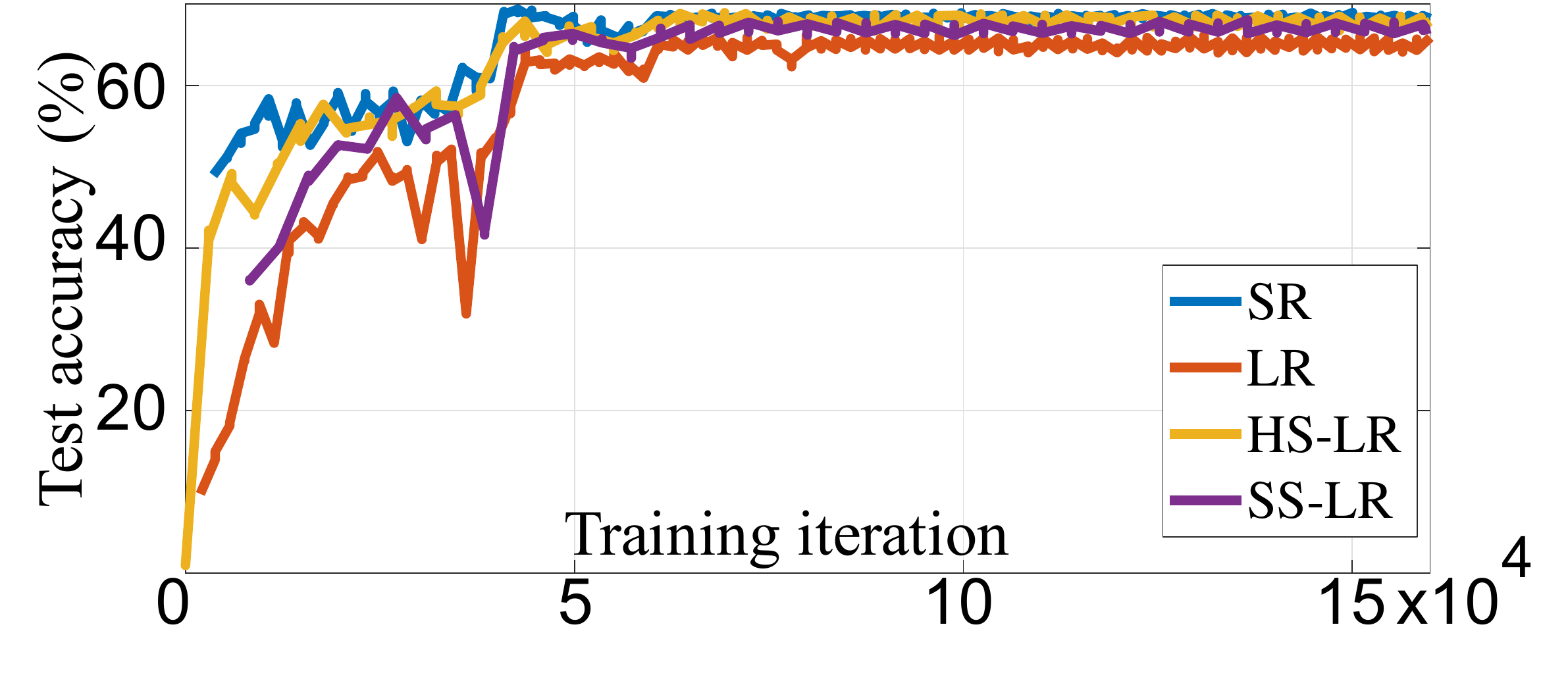}
	\captionof{figure}{Evaluating the converging rate on CIFAR100.}
	\label{fig:test1}
\end{figure}

\noindent{\bf Convergence Rate. }
Figure \ref{fig:test1} compares the convergence rate of 
SR and LR  
on CIFAR100.
It is shown that 
all learning algorithms have very similar convergence speeds,
suggesting that our method does not sacrifice the training efficiency
whilst yielding favourable performance advantages.

\begin{figure}[th]
	\centering
	\includegraphics[width=0.9\linewidth]{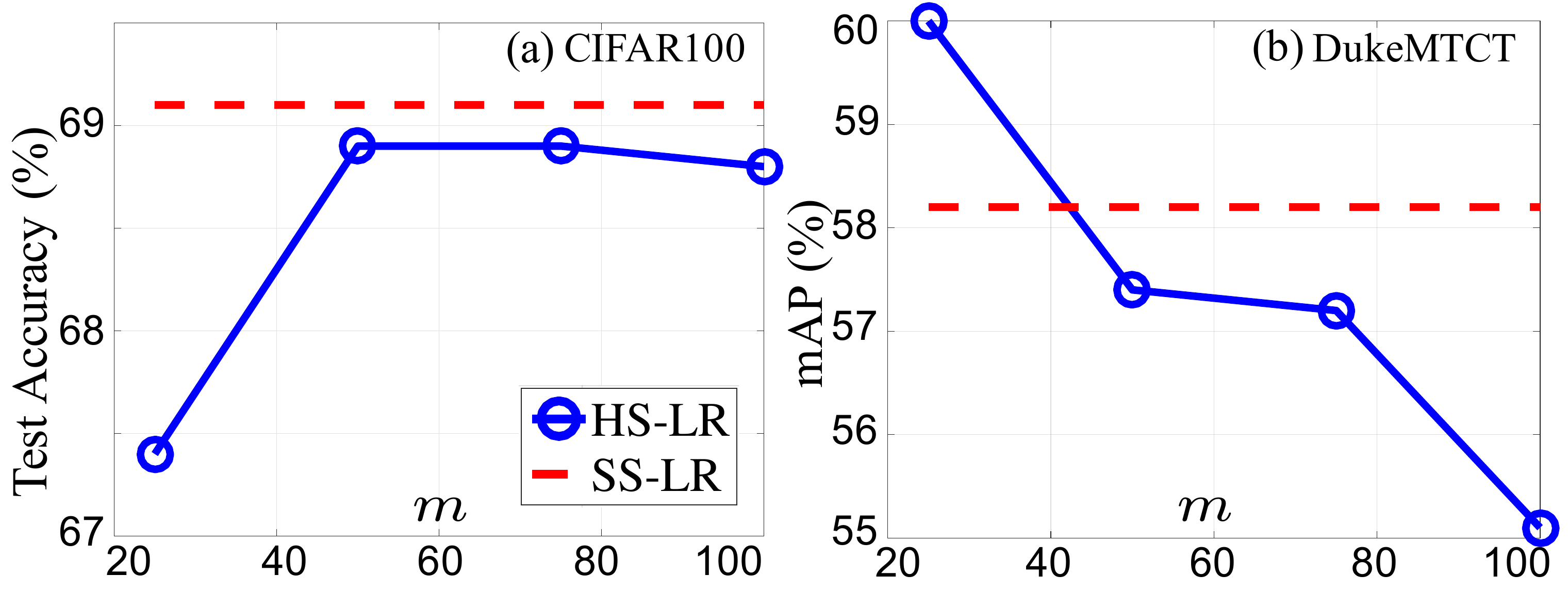}
	\captionof{figure}{
		Sensitivity evaluation of $m$ in HS-LR.
	}
	\label{fig:HSvsSS}
\end{figure}

\begin{figure}[h]
	\centering
	\includegraphics[width=0.6\columnwidth]{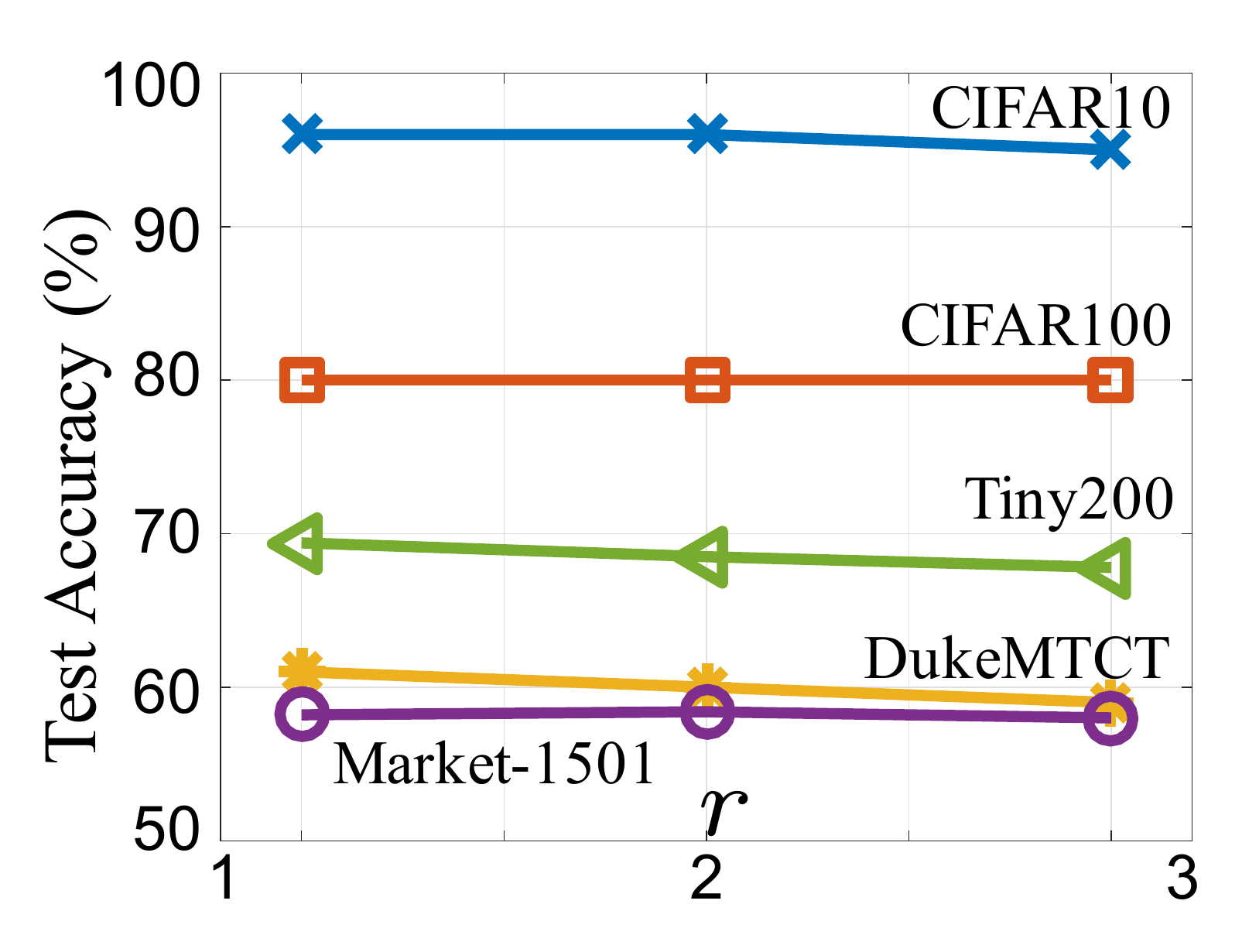}
	\captionof{figure}{Sensitivity evaluation of $\gamma$ in SS-LR.}
	\label{fig:gamma}
\end{figure}

\noindent{\bf Parameter Analysis. }
We analysed the parameter sensitivity of 
HS ($m$ in Eq \eqref{eq:LR_sel}) and
SS ($r$ in Eq \eqref{eq:LR_att}) designs.
Fig \ref{fig:HSvsSS} shows that
a good selection of $m$ is important, and
a high value of $m$ is preferred for object classification
but hurts the performance of person re-id.
This is consistent with the earlier observation
that re-id needs to mine fine-grained discriminative information
by concentrating the learning attention more on the most confusing negative classes.
As shown in Fig \ref{fig:gamma},
$r$ is not sensitive to the model performance ($r\!=\!2$ in the main experiments),
rendering SS a favourable choice over HS
in terms of parameter selection.

\noindent{\bf Hard Mining for SR. }
How is the performance of the proposed focus rectified hard mining on SR? We additionally applied the same HS formulation (Eq \eqref{eq:LR_sel}) to SR, 
and tested two cases: (1) With MobileNet on DukeMTMC, we obtained 54.9\%/34.9\% Rank-1/mAP vs 57.0\%/35.8\% by the standard SR. (2) With WRN-28-10
on CIFAR100, no performance change, both at 80.0\%. This suggests SR does not suffer from the same NCD problem as LR.

\section{Conclusion}
In this work, we have extensively investigated the validity and advantages of
the logistic regression (LR) learning algorithms
for training single-label multi-class neural network classifiers,
a standard technique conventionally employed for multi-label classification model learning. 
This is motivated by our in-depth analyses of softmax regression (SR) and LR in learning properties and their correlation.
We identified the negative class distraction problem 
and proposed two rectification solutions using a hard mining idea.
Extensive experiments on both 
coarse-grained object classification
and fine-grained person re-identification and spare attribute recognition tasks
show the performance effectiveness of the proposed LR algorithms 
over the standard choice SR. 

\section{Acknowledgements}
This work was partly supported by the China Scholarship Council, Vision Semantics Limited, the Royal Society Newton Advanced Fellowship Programme (NA150459), and Innovate UK Industrial
Challenge Project on Developing and Commercialising Intelligent Video Analytics Solutions for
Public Safety (98111-571149). 
\bibliographystyle{aaai}
\bibliography{reference}

\begin{thebibliography}{}

\bibitem[\protect\citeauthoryear{Abadi \bgroup et al\mbox.\egroup
  }{2016}]{abadi2016tensorflow}
Abadi, M.; Barham, P.; Chen, J.; Chen, Z.; Davis, A.; Dean, J.; Devin, M.;
  Ghemawat, S.; Irving, G.; Isard, M.; et~al.
\newblock 2016.
\newblock Tensorflow: A system for large-scale machine learning.
\newblock In {\em OSDI}.

\bibitem[\protect\citeauthoryear{Akbani, Kwek, and
  Japkowicz}{2004}]{akbani2004applying}
Akbani, R.; Kwek, S.; and Japkowicz, N.
\newblock 2004.
\newblock Applying support vector machines to imbalanced datasets.
\newblock In {\em ECML}.

\bibitem[\protect\citeauthoryear{Belkin, Niyogi, and
  Sindhwani}{2006}]{belkin2006manifold}
Belkin, M.; Niyogi, P.; and Sindhwani, V.
\newblock 2006.
\newblock Manifold regularization: A geometric framework for learning from
  labeled and unlabeled examples.
\newblock {\em JMLR} 7(Nov):2399--2434.

\bibitem[\protect\citeauthoryear{Bishop}{2006}]{Bishop}
Bishop, C.~M.
\newblock 2006.
\newblock Pattern recognition and machine learning (information science and
  statistics).

\bibitem[\protect\citeauthoryear{Bridle}{1990}]{bridle1990probabilistic}
Bridle, J.~S.
\newblock 1990.
\newblock Probabilistic interpretation of feedforward classification network
  outputs, with relationships to statistical pattern recognition.
\newblock In {\em Neurocomputing}.

\bibitem[\protect\citeauthoryear{Chaudhari \bgroup et al\mbox.\egroup
  }{2016}]{chaudhari2016entropy}
Chaudhari, P.; Choromanska, A.; Soatto, S.; LeCun, Y.; Baldassi, C.; Borgs, C.;
  Chayes, J.; Sagun, L.; and Zecchina, R.
\newblock 2016.
\newblock Entropy-sgd: Biasing gradient descent into wide valleys.
\newblock In {\em ICLR}.

\bibitem[\protect\citeauthoryear{Chua \bgroup et al\mbox.\egroup
  }{2009}]{chua2009nus}
Chua, T.-S.; Tang, J.; Hong, R.; Li, H.; Luo, Z.; and Zheng, Y.
\newblock 2009.
\newblock Nus-wide: a real-world web image database from national university of
  singapore.
\newblock In {\em CIVR}.

\bibitem[\protect\citeauthoryear{Deng \bgroup et al\mbox.\egroup
  }{2009}]{deng2009imagenet}
Deng, J.; Dong, W.; Socher, R.; Li, L.-J.; Li, K.; and Fei-Fei, L.
\newblock 2009.
\newblock Imagenet: A large-scale hierarchical image database.
\newblock In {\em CVPR}.

\bibitem[\protect\citeauthoryear{Dong, Gong, and Zhu}{2017a}]{dong2017class}
Dong, Q.; Gong, S.; and Zhu, X.
\newblock 2017a.
\newblock Class rectification hard mining for imbalanced deep learning.
\newblock In {\em Proceedings of the IEEE International Conference on Computer
  Vision}.

\bibitem[\protect\citeauthoryear{Dong, Gong, and Zhu}{2017b}]{dong2017multi}
Dong, Q.; Gong, S.; and Zhu, X.
\newblock 2017b.
\newblock Multi-task curriculum transfer deep learning of clothing attributes.
\newblock In {\em 2017 IEEE Winter Conference on Applications of Computer
  Vision (WACV)}.

\bibitem[\protect\citeauthoryear{Dong, Gong, and
  Zhu}{2018}]{dong2018imbalanced}
Dong, Q.; Gong, S.; and Zhu, X.
\newblock 2018.
\newblock Imbalanced deep learning by minority class incremental rectification.
\newblock {\em IEEE transactions on pattern analysis and machine intelligence}.

\bibitem[\protect\citeauthoryear{Goodfellow, Bengio, and
  Courville}{2016}]{goodfellow2016deep}
Goodfellow, I.; Bengio, Y.; and Courville, A.
\newblock 2016.
\newblock {\em Deep learning}.
\newblock MIT press Cambridge.

\bibitem[\protect\citeauthoryear{He and Garcia}{2009}]{he2009learning}
He, H., and Garcia, E.~A.
\newblock 2009.
\newblock Learning from imbalanced data.
\newblock {\em TKDE}.

\bibitem[\protect\citeauthoryear{He \bgroup et al\mbox.\egroup
  }{2016}]{he2016deep}
He, K.; Zhang, X.; Ren, S.; and Sun, J.
\newblock 2016.
\newblock Deep residual learning for image recognition.
\newblock In {\em CVPR}.

\bibitem[\protect\citeauthoryear{Heinze and
  Schemper}{2002}]{heinze2002solution}
Heinze, G., and Schemper, M.
\newblock 2002.
\newblock A solution to the problem of separation in logistic regression.
\newblock {\em Statistics in medicine}.

\bibitem[\protect\citeauthoryear{Howard \bgroup et al\mbox.\egroup
  }{2017}]{howard2017mobilenets}
Howard, A.~G.; Zhu, M.; Chen, B.; Kalenichenko, D.; Wang, W.; Weyand, T.;
  Andreetto, M.; and Adam, H.
\newblock 2017.
\newblock Mobilenets: Efficient convolutional neural networks for mobile vision
  applications.
\newblock {\em arXiv preprint arXiv:1704.04861}.

\bibitem[\protect\citeauthoryear{Huang \bgroup et al\mbox.\egroup
  }{2016}]{huang2016learning}
Huang, C.; Li, Y.; Change~Loy, C.; and Tang, X.
\newblock 2016.
\newblock Learning deep representation for imbalanced classification.
\newblock In {\em CVPR}.

\bibitem[\protect\citeauthoryear{Huang \bgroup et al\mbox.\egroup
  }{2017}]{huang2017densely}
Huang, G.; Liu, Z.; Van Der~Maaten, L.; and Weinberger, K.~Q.
\newblock 2017.
\newblock Densely connected convolutional networks.
\newblock In {\em CVPR}.

\bibitem[\protect\citeauthoryear{Japkowicz and
  Stephen}{2002}]{japkowicz2002class}
Japkowicz, N., and Stephen, S.
\newblock 2002.
\newblock The class imbalance problem: A systematic study.
\newblock {\em IDA}.

\bibitem[\protect\citeauthoryear{King and Zeng}{2001}]{king2001logistic}
King, G., and Zeng, L.
\newblock 2001.
\newblock Logistic regression in rare events data.
\newblock {\em Political analysis}.

\bibitem[\protect\citeauthoryear{Kingma and Ba}{2015}]{kingma2014adam}
Kingma, D., and Ba, J.
\newblock 2015.
\newblock Adam: A method for stochastic optimization.
\newblock In {\em ICLR}.

\bibitem[\protect\citeauthoryear{Krishnapuram \bgroup et al\mbox.\egroup
  }{2005}]{krishnapuram2005sparse}
Krishnapuram, B.; Carin, L.; Figueiredo, M.~A.; and Hartemink, A.~J.
\newblock 2005.
\newblock Sparse multinomial logistic regression: Fast algorithms and
  generalization bounds.
\newblock {\em TPAMI}.

\bibitem[\protect\citeauthoryear{Krizhevsky and
  Hinton}{2009}]{krizhevsky2009learning}
Krizhevsky, A., and Hinton, G.
\newblock 2009.
\newblock Learning multiple layers of features from tiny images.
\newblock {\em Technical report, University of Toronto}.

\bibitem[\protect\citeauthoryear{Krizhevsky, Sutskever, and
  Hinton}{2012}]{krizhevsky2012imagenet}
Krizhevsky, A.; Sutskever, I.; and Hinton, G.~E.
\newblock 2012.
\newblock Imagenet classification with deep convolutional neural networks.
\newblock In {\em NIPS}.

\bibitem[\protect\citeauthoryear{LeCun \bgroup et al\mbox.\egroup
  }{1989}]{lecun1989backpropagation}
LeCun, Y.; Boser, B.; Denker, J.~S.; Henderson, D.; Howard, R.~E.; Hubbard, W.;
  and Jackel, L.~D.
\newblock 1989.
\newblock Backpropagation applied to handwritten zip code recognition.
\newblock {\em NC}.

\bibitem[\protect\citeauthoryear{Lin \bgroup et al\mbox.\egroup
  }{2017}]{lin2017focal}
Lin, T.-Y.; Goyal, P.; Girshick, R.; He, K.; and Doll.
\newblock 2017.
\newblock Focal loss for dense object detection.
\newblock In {\em ICCV}.

\bibitem[\protect\citeauthoryear{Lin, Chen, and Yan}{2014}]{lin2013network}
Lin, M.; Chen, Q.; and Yan, S.
\newblock 2014.
\newblock Network in network.
\newblock In {\em ICLR}.

\bibitem[\protect\citeauthoryear{Little}{1974}]{little1974existence}
Little, W.~A.
\newblock 1974.
\newblock The existence of persistent states in the brain.
\newblock In {\em From High-Temperature Superconductivity to Microminiature
  Refrigeration}.

\bibitem[\protect\citeauthoryear{Liu \bgroup et al\mbox.\egroup
  }{2016}]{liu2016deepfashion}
Liu, Z.; Luo, P.; Qiu, S.; Wang, X.; and Tang, X.
\newblock 2016.
\newblock Deepfashion: Powering robust clothes recognition and retrieval with
  rich annotations.
\newblock In {\em CVPR}.

\bibitem[\protect\citeauthoryear{Luce}{2005}]{luce2005individual}
Luce, R.~D.
\newblock 2005.
\newblock {\em Individual choice behavior: A theoretical analysis}.

\bibitem[\protect\citeauthoryear{Maas, Hannun, and Ng}{}]{maas2013rectifier}
Maas, A.~L.; Hannun, A.~Y.; and Ng, A.~Y.
\newblock Rectifier nonlinearities improve neural network acoustic models.
\newblock In {\em ICML}.

\bibitem[\protect\citeauthoryear{Mor-Yosef \bgroup et al\mbox.\egroup
  }{1990}]{mor1990ranking}
Mor-Yosef, S.; Samueloff, A.; Modan, B.; Navot, D.; and Schenker, J.~G.
\newblock 1990.
\newblock Ranking the risk factors for cesarean: logistic regression analysis
  of a nationwide study.
\newblock {\em Obstetrics and gynecology}.

\bibitem[\protect\citeauthoryear{Peterson and
  S{\"o}derberg}{1989}]{peterson1989new}
Peterson, C., and S{\"o}derberg, B.
\newblock 1989.
\newblock A new method for mapping optimization problems onto neural networks.
\newblock {\em IJNS}.

\bibitem[\protect\citeauthoryear{Qiu \bgroup et al\mbox.\egroup
  }{2013}]{qiu2013logistic}
Qiu, Z.; Li, H.; Su, H.; Ou, G.; and Wang, T.
\newblock 2013.
\newblock Logistic regression bias correction for large scale data with rare
  events.
\newblock In {\em ICADMA}.

\bibitem[\protect\citeauthoryear{Ristani \bgroup et al\mbox.\egroup
  }{2016}]{ristani2016MTMC}
Ristani, E.; Solera, F.; Zou, R.; Cucchiara, R.; and Tomasi, C.
\newblock 2016.
\newblock Performance measures and a data set for multi-target, multi-camera
  tracking.
\newblock In {\em ECCV workshop on Benchmarking Multi-Target Tracking}.

\bibitem[\protect\citeauthoryear{Russakovsky \bgroup et al\mbox.\egroup
  }{2015}]{russakovsky2015imagenet}
Russakovsky, O.; Deng, J.; Su, H.; Krause, J.; Satheesh, S.; Ma, S.; Huang, Z.;
  Karpathy, A.; Khosla, A.; Bernstein, M.; et~al.
\newblock 2015.
\newblock Imagenet large scale visual recognition challenge.
\newblock {\em IJCV}.

\bibitem[\protect\citeauthoryear{Schaefer}{1983}]{schaefer1983bias}
Schaefer, R.~L.
\newblock 1983.
\newblock Bias correction in maximum likelihood logistic regression.
\newblock {\em Statistics in Medicine}.

\bibitem[\protect\citeauthoryear{Srivastava \bgroup et al\mbox.\egroup
  }{2014}]{srivastava2014dropout}
Srivastava, N.; Hinton, G.~E.; Krizhevsky, A.; Sutskever, I.; and
  Salakhutdinov, R.
\newblock 2014.
\newblock Dropout: a simple way to prevent neural networks from overfitting.
\newblock {\em JMLR}.

\bibitem[\protect\citeauthoryear{Weiss}{2004}]{weiss2004mining}
Weiss, G.~M.
\newblock 2004.
\newblock Mining with rarity: a unifying framework.
\newblock {\em ACM SIGKDD Explorations Newsletter}.

\bibitem[\protect\citeauthoryear{Zagoruyko}{2016}]{zagoruyko2016wide}
Zagoruyko, S. e.~a.
\newblock 2016.
\newblock Wide residual networks.
\newblock {\em arXiv preprint arXiv:1605.07146}.

\bibitem[\protect\citeauthoryear{Zheng \bgroup et al\mbox.\egroup
  }{2015}]{zheng2015scalable}
Zheng, L.; Shen, L.; Tian, L.; Wang, S.; Wang, J.; and Tian, Q.
\newblock 2015.
\newblock Scalable person re-identification: A benchmark.
\newblock In {\em ICCV}.

\bibitem[\protect\citeauthoryear{Zheng, Zheng, and
  Yang}{2017}]{zheng2017unlabeled}
Zheng, Z.; Zheng, L.; and Yang, Y.
\newblock 2017.
\newblock Unlabeled samples generated by gan improve the person
  re-identification baseline in vitro.
\newblock In {\em ICCV}.

\end{thebibliography}
\end{document}